CHAPTER TWO

ENHANCING MACHINE LEARNING
MODEL EFFICIENCY THROUGH QUANTIZATION
AND BIT DEPTH OPTIMIZATION:
A PERFORMANCE ANALYSIS
ON HEALTHCARE DATA

MITUL GOSWAMI AND ROMIT CHATTERJEE

**Abstract:** This research aims to optimize intricate learning models by implementing quantization and bit-depth optimization techniques. The objective is to significantly cut time complexity while preserving model efficiency, thus addressing the challenge of extended execution times in intricate models. Two medical datasets were utilized as case studies to apply a Logistic Regression (LR) machine learning model. Using efficient quantization and bit depth optimization strategies the input data is downscaled from float64 to float32 and int32. The results demonstrated a significant reduction in time complexity, with only a minimal decrease in model accuracy post-optimization, showcasing the state-of-the-art optimization approach. This comprehensive study concludes that the impact of these optimization techniques varies depending on a set of parameters.

**Keywords:** Bit-Depth Optimization, Time Complexity, Logistic Regression, Quantization.

## Introduction

Machine learning, an important aspect of artificial intelligence, allows computers to learn from experience without being explicitly programmed. Machine learning algorithms use large datasets to find patterns and make data-driven predictions, affecting fields such as image recognition, language translation, and more [1]. The models include supervised, unsupervised, and reinforcement learning, which drive innovation across diverse sectors.

In recent years, there have been remarkable strides in optimization approaches for machine learning models, which have garnered significant research attention. Innovations such as adaptive learning rates have been sightseen to enhance the efficiency of model training and convergence [2]. To address the challenges associated with training large-scale models, researchers have developed distributed and parallel processing techniques that significantly reduce training time [3]. Moreover, robust optimization techniques have been introduced to manage noisy data, ensuring safer and more reliable deployment in real-world scenarios [4]. These advancements not only accelerate the development of machine learning models but also enhance their accuracy and robustness, making them more applicable across various sectors. As the field continues to evolve, ongoing research is expected to yield even more sophisticated and efficient optimization methods [5].

Logistic Regression is a fundamental machine learning model used for binary classification tasks. It predicts the probability that a given input belongs to a certain class by applying a logistic function to a linear combination of input features. The model outputs values between 0 and 1, which are interpreted as probabilities [6]. Logistic Regression is valued for its simplicity, interpretability, and efficiency, especially in scenarios where the relationship between the independent variables and the target is approximately linear. It's widely used in various fields, including medical diagnosis, finance, and social sciences, for tasks like disease prediction and risk assessment [7]. Medical and healthcare datasets play a vital role in driving progress and innovation within machine learning (ML). Comprising comprehensive patient information, medical imagery, genomic data, and clinical trial results, these datasets provide the foundation for building sophisticated ML models capable of revolutionizing healthcare services [8]. Through the use of healthcare datasets, researchers are able to develop predictive models aimed at early disease detection, and optimize therapeutic approaches [9][10].

# Related Works

In recent years, advancements in the application of optimization techniques, particularly using quantization, have significantly enhanced the performance and efficiency of ML models for various applications and use cases. S. Sun et al., outlined the machine learning optimization difficulties and provided an overview of the fundamentals and developments of popular optimization techniques [11]. The authors also looked at a few difficulties and unsolved issues with machine learning optimization. In a different perspective, L. Yang et al., optimized the hyperparameters of common machine learning models, and various state-of-the-art optimization techniques were introduced and discussed for their application to machine learning algorithms [12]. Additionally, experiments were conducted on benchmark datasets to compare the performance of different optimization methods, providing practical examples of hyperparameter optimization. Similarly, K.M. Hamdia et al., proposed an improved Deep Neural Network (DNN) model that outperformed the traditional one-hidden layer network in terms of prediction accuracy. Additionally, the model fared better in GA than ANFIS with a much smaller number of generations [13]. J. W. La et al., suggested a Bayesian optimization-based technique that, when time cost is considered, can determine the optimal hyperparameters for popular machine learning models, including neural networks, random forests, and even multi-grained cascade forests [14]. Similarly, M. Fairley et al., developed a machine learning and generalizable optimization strategy to optimize the order of operating room operations and reduce delays resulting from PACU unavailability [15].

O. Hrizi et al., presented a machine learning-based multi-task optimized model that chooses the classifiers' hyper-parameters and extracts the best texture features from TB-related photos. minimizing the amount of features collected while raising the accuracy rate [16]. Y. Choukroun et al., suggested studying and improving limited MSE issues for effective hardware-aware quantization. The suggested method enables pre-trained models to be deployed on constrained hardware resources by allowing 4-bit integer (INT4) quantization [17]. Furthermore, B. Rokh et al., provided a thorough analysis of quantization methods and techniques, emphasizing image categorization. The authors studied the use of a scale factor parameter for full-precision value approximation and developed quantization techniques based on clustering [18].

While modern, state-of-the-art optimization techniques are adaptive and cater to various applications, they may not be ideally suited for medical datasets. Our study presents a performance-based comparative analysis, focusing on the fusion of quantization and bit-depth optimization specifically tailored to Logistic Regression models applied to medical datasets [19]. Complex and non-uniform distributions are common in medical datasets; excessive values may indicate uncommon medical illnesses, patient outliers, or particular clinical occurrences. Such complexity is ideally suited for QuantileTransformer to handle. Transforming data into a uniform distribution guarantee that there are about equal numbers of data points in each quantile. Medical datasets are known for their high values, which are mitigated by an even distribution. Numpy.round is a useful tool for quantizing medical data because of its straightforward and effective method of rounding numerical data. Maintaining the highest level of precision in data may not always be necessary in the healthcare industry. KBinsDiscretizer is tailored to convert unbroken data into discrete periods, making it particularly useful for medical datasets with a wide range of features [20].

## Quantization and Bit-Depth Optimization

The medical datasets are quantized using the QuantileTransformer function, which is used in the optimization of machine learning models such as Logistic Regression (LR). Bit representation of the data is reduced by bit depth optimization and quantization. An array named $X_{Quantized}$ holds the converted data. The quantized data is then transformed into float32 and int32 forms by using the astype() method to convert it from 64-bit to 32-bit. The mathematical process for applying quantization with QuantileTransformer is summarized below:

$$X_{quantized} = Q(x) = q \times \frac{X - P_{min}}{P_{max} - P_{min}} \tag{1}$$

Equation (1) quantizes the dataset X by normalizing its values to a [0, 1] range using $\frac{X - P_{min}}{P_{max} - P_{min}}$. The normalized data is then multiplied by a vector of quantiles q calculated by the QuantileTransformer, mapping each value to its corresponding quantile-based representation. This process ensures that the transformed data $X_{Quantized}$, follows a consistent distribution, reducing the impact of outliers and skewness. Quantization enhances model robustness by standardizing feature ranges across the dataset.

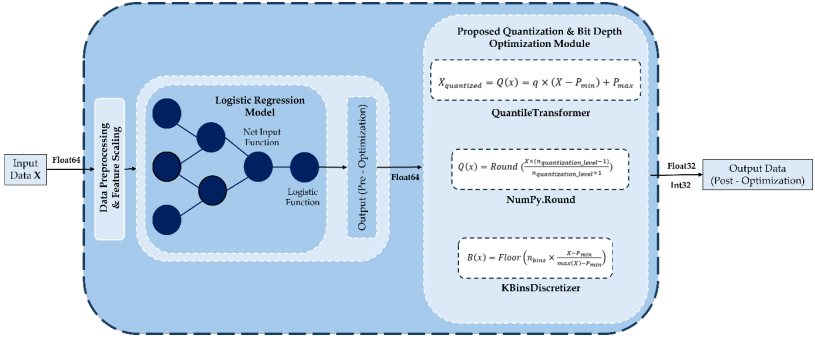

**Fig. 1.** Model Workflow Diagram

In a similar approach, the Numpy.round function is utilized for quantization within the LR model on datasets. The Numpy.round function specifically rounds the values in $X_{train}$ and $X_{test}$ to four decimal places, reducing the data precision from 64 bits to 12 bits. The rounded values are subsequently stored in $X_{train\_quantized}$ and $X_{test\_quantized}$.

$$Q(x) = Round\left(\frac{(X-P_{min}) \times (n_{levels}-1)}{P_{max}-P_{min}}\right) \qquad (2)$$

Equation (2) quantizes the dataset $X$ by first normalizing each value to the range [0, 1] using the round function. The normalized data is then scaled, which determines the granularity of quantization. The Round function maps each value to the nearest discrete level, producing the quantized output $Q(X)$. This approach ensures uniform scaling, enabling effective data compression and reducing precision while preserving essential patterns for downstream tasks.

The KBinsDiscretizer function is applied to purposefully lower the bit precision of the data representation, improving the performance of the LR model. By using this function, the input data is quantized. The equation for quantization using the KBinsDiscretizer function is outlined below:

$$B(x) = Floor\left(\frac{n_{bins} \times (X-P_{min})}{P_{max}-P_{min}}\right) \qquad (3)$$

In equation (3), X denotes the input data with dimensions ($n_{samples}$, $n_{features}$), while $n_{bins}$ represents the number of bins for quantization. This equation bins the dataset by normalizing each value between the minimum and maximum range using $\frac{X - P_{min}}{P_{max} - P_{min}}$. The normalized value is then multiplied by $n_{bins}$, representing the number of discrete bins. The Floor function maps each value to the nearest lower bin, ensuring consistent binning. This method is commonly used for data discretization, transforming continuous data into categorical bins, facilitating easier pattern recognition while maintaining essential distribution characteristics.

# Experimentation and Results

As part of the model training, the optimization techniques conferred above were applied and tested on two medical datasets using the Logistic Regression (LR) model. By employing these optimization methods, the bit precision of the input data was reduced.

**Heart Disease Prediction**

The dataset includes various medical details of individuals, such as age, gender, type of chest pain, resting blood pressure, level of cholesterol, fasting blood sugar, resting electrocardiogram results, maximum heart rate achieved during exercise, exercise-induced angina, ST depression due to exercise compared to rest, and the slope of the peak exercise ST segment. The target variable (target) indicates the presence (1) or absence (0) of cardiac disease. This dataset is likely used for categorizing heart disease or assessing risk. The next section delves into the application of machine learning models on this data.

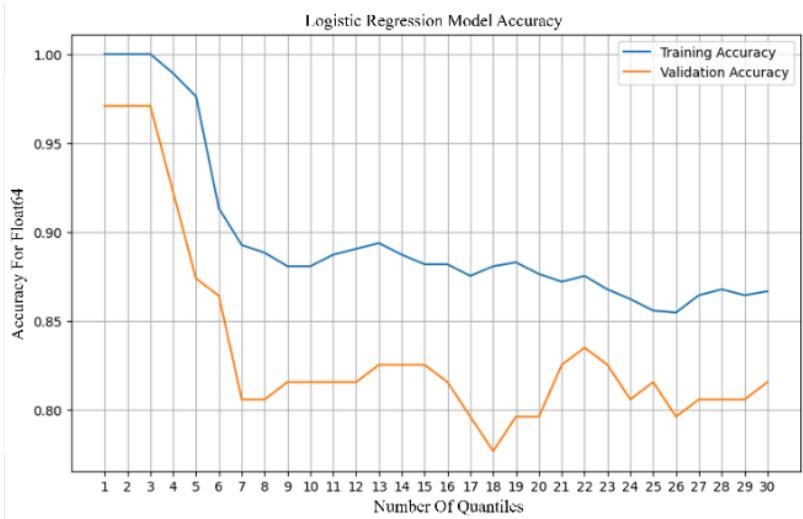

**Fig.2**. Model Accuracy of Logistic Regression Model Pre-Optimization

The dataset is loaded for regression using the Logistic Regression (LR) model. Features are normalized using StandardScaler, and the data is split into 90% training and 10% testing sets. The LR model is trained with the default k value (typically 5), and predictions are made on the test set. The detailed results of the model have been provided in Table. 1. Fig. 2 illustrates the model's accuracy to the number of quantiles.

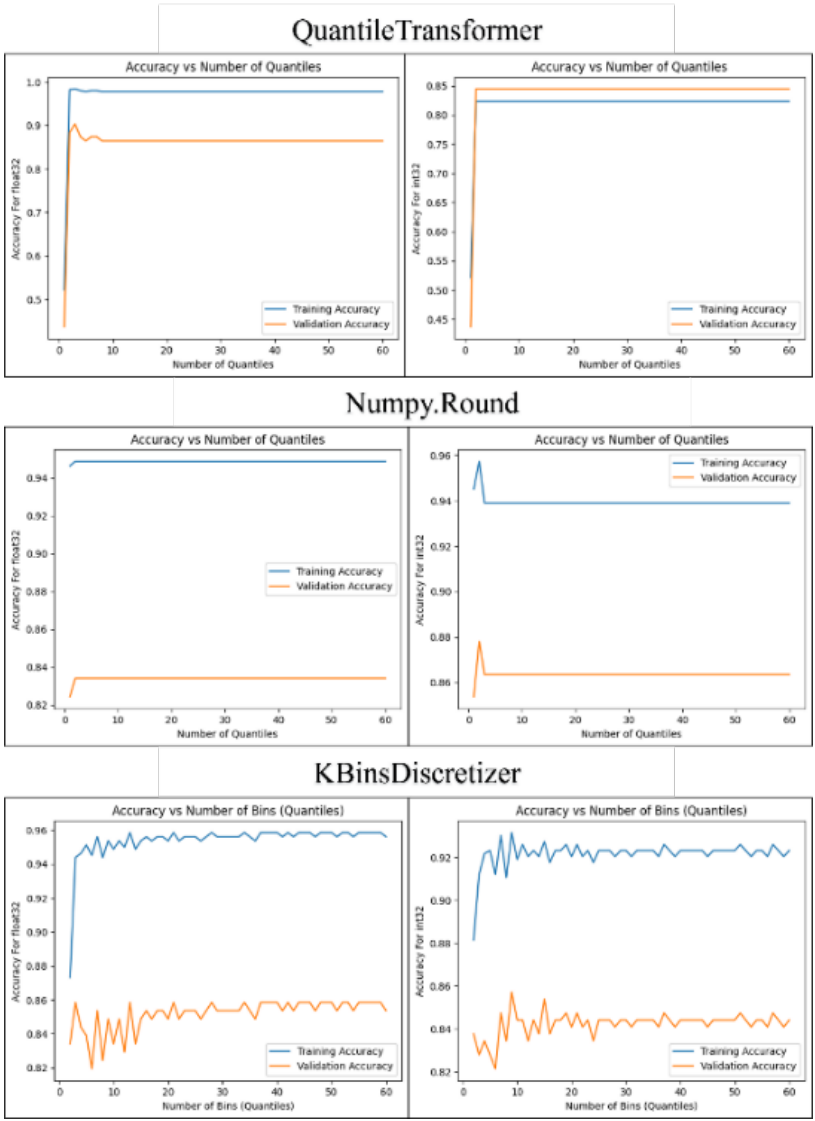

**Fig. 3**. Logistic Regression Model Performance Post-Optimization

When optimization is applied using the QuantileTransformer, the results of the quantization techniques are meticulously represented in Table. 1. Fig. 3 showcases the model's performance after optimization across all techniques.

## Breast Cancer Detection

The dataset serves as a vital tool for healthcare research, containing key information derived from breast cancer images. It includes important characteristics of the masses and corresponding labels that indicate whether they are malignant. Early detection plays a critical role in improving treatment options and outcomes. Additionally, the dataset facilitates the development of computer-aided diagnostic systems, enhancing diagnostic accuracy and supporting personalized care. The following section discusses the use of the LR model, along with the results obtained.

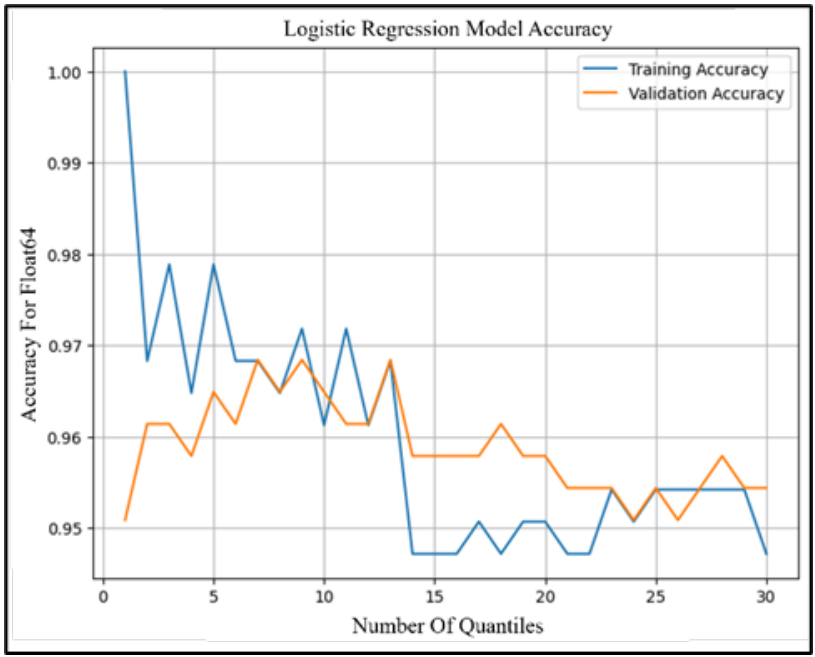

**Fig. 4**. Model Accuracy of Logistic Regression Model Pre-Optimization

Logistic Regression (LR) is employed to classify breast tumor experiments. Scaled features are essential for regression algorithms like LR. The LR model is trained on this data and used to predict labels for the test set, with accuracy assessed accordingly. Model efficiency is measured by the time complexity of the training process. Fig. 4 illustrates accuracy as a function of the number of quantiles.

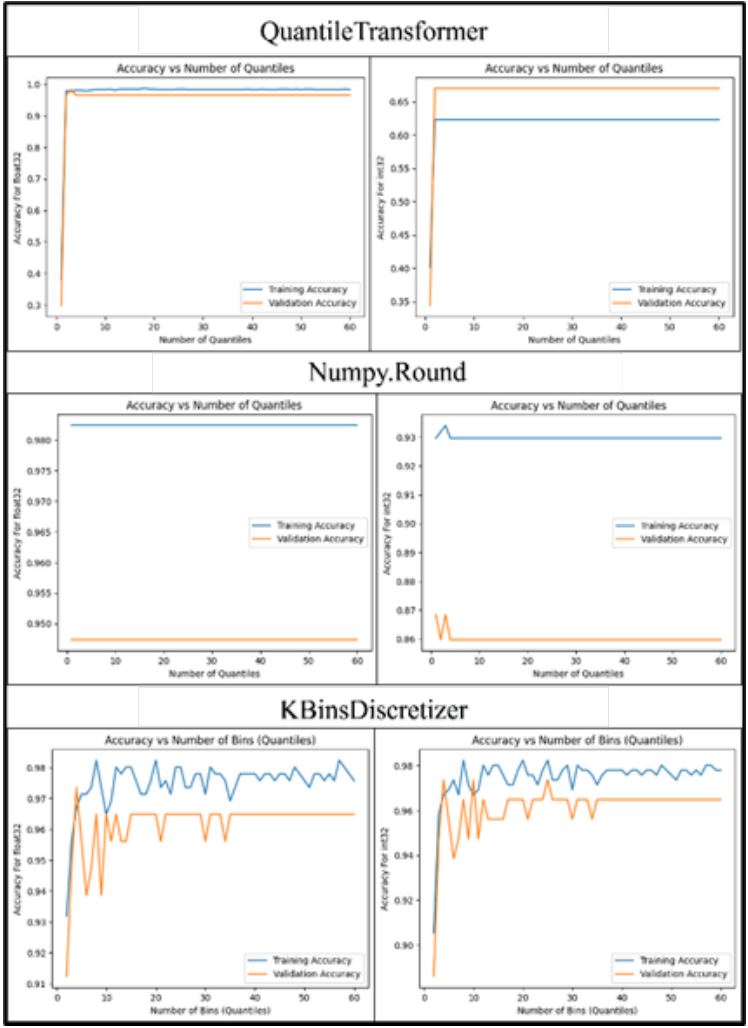

**Fig. 5.** Logistic Regression Model Performance Post-Optimization

After optimization, the model's results are detailed in Table 1, providing a comprehensive comparison of key performance metrics. Figure 5 visually illustrates the enhanced performance across various optimization techniques, highlighting improvements in accuracy, efficiency, and robustness. This analysis demonstrates the effectiveness of the applied optimization strategies in refining model outputs.

# Discussion and Conclusion

The evaluation highlights the impact of optimization on both performance metrics and computational efficiency, demonstrating the effectiveness of various approaches in refining the model. For the Heart Disease Prediction Dataset, QuantileTransformer demonstrates the greatest consistency in maintaining model accuracy. However, Numpy.round and KBinsDiscretizer result in minor accuracy reductions. Of these methods, KBinsDiscretizer achieves the best accuracy, followed closely by Numpy.round and QuantileTransformer. Table. 1 and Fig. 6 provide a detailed comparison of the study.

**Table. 1.** Logistic Regression Model Findings

| Dataset Name | Optimizing Technique | Model Accuracy (%) | | | Time Complexity (Sec) | | |
|---|---|---|---|---|---|---|---|
| | | *Before Optimization* | *Float 32* | *Int32* | *Before Optimization* | *Float 32* | *Int32* |
| Heart Disease Prediction | QuantileTransfomrer | 87.38 | 86.41 | 84.47 | 0.0029 | 0.0023 | 0.0014 |
| | Numpy.round | | 83.41 | 81.27 | | 0.0024 | 0.0011 |
| | KBinsDiscretizer | | 85.37 | 84.42 | | 0.0021 | 0.0017 |
| Breast Cancer Detection | QuantileTransfomrer | 96.49 | 95.18 | 67.74 | 0.0258 | 0.0142 | 0.0025 |
| | Numpy.round | | 96.05 | 88.16 | | 0.0068 | 0.0027 |
| | KBinsDiscretizer | | 96.14 | 94.99 | | 0.0073 | 0.0017 |

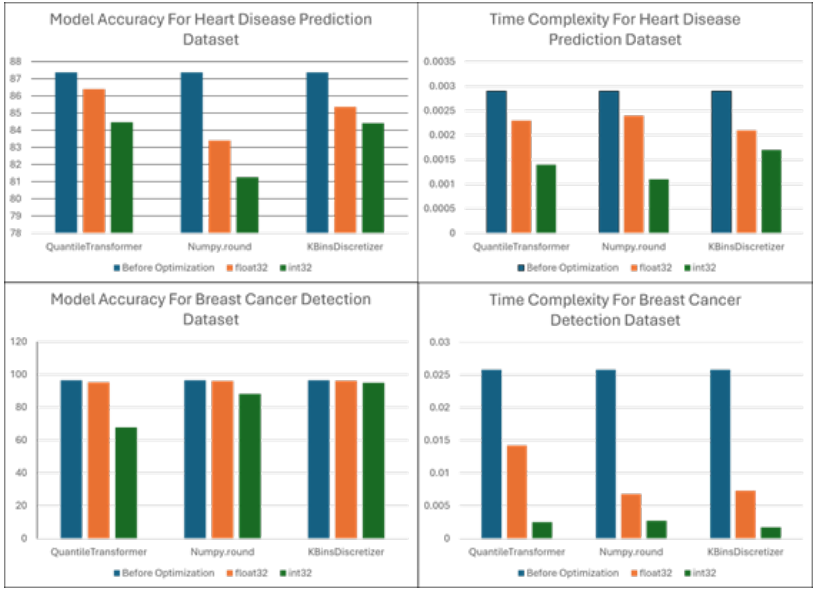

**Fig 6.** Comparative Chart for Findings of the Logistic Regression Model

For the Breast Cancer Detection Dataset, applying QuantileTransformer quantization results in only a 1.31% accuracy decrease compared to the non-optimized model when using float32 bit reduction, while also achieving a significant 92.1% decrease in time complexitySSSS. In comparison, QuantileTransformer is the least effective, as its significant accuracy loss diminishes the practicality of the optimization.

The results demonstrate that quantization is a valuable optimization technique for reducing the time complexity of machine learning models with minimal impact on accuracy. However, the choice of the quantization method is critical, as some methods are more effective than others in lowering time complexity while preserving accuracy. It's also noteworthy that the time complexity evaluations were performed using a single CPU core. In real-world applications, time complexity could be further reduced by leveraging multiple CPU cores or GPUs [23][24]. In summary, this study shows that quantization can greatly improve machine learning model performance when applied to medical datasets. The extensive experiments reveal that the efficiency of optimization techniques varies depending on the model and dataset in use. Therefore, the success of optimization hinges not just on the method itself but also on the specific model and dataset involved.